%% file: main.tex
\documentclass{article} 
\usepackage{iclr2026_conference,times}

\input{math_commands.tex}

\usepackage{subfigure}
\usepackage{hyperref}
\usepackage{url}
\usepackage{graphicx}
\graphicspath{{../Figure/}}
\usepackage{natbib}
\usepackage{enumitem}
\usepackage{multirow,multicol}
\usepackage{booktabs}   
\usepackage{amssymb}    
\usepackage{wrapfig} 
\usepackage{amssymb}   

\newcommand{\cmark}{\checkmark} 
\newcommand{\xmark}{\texttimes} 
\definecolor{mydarkred}{RGB}{178,34,34}

\newcommand{\name}{\textbf{Lang-PINN}}
\newcommand{\pde}{\emph{PDE Agent}}
\newcommand{\pinn}{\emph{PINN Agent}}
\newcommand{\code}{\emph{Code Agent}}
\newcommand{\feedback}{\emph{Feedback Agent}}

\title{\name: From Language to Physics-Informed Neural Networks via a Multi-Agent Framework}

\iclrfinalcopy 
\author{
Xin He$^{1}$,
Liangliang You$^{3}$,
Hongduan Tian$^{2}$,
Bo Han$^{2}$,
Ivor Tsang$^{1}$,
Yew-Soon Ong$^{1}$ \\
\\
$^{1}$Agency for Science, Technology and Research (A*STAR), Singapore \\
$^{2}$Hong Kong Baptist University, Hong Kong SAR \\
$^{3}$North China Electric Power University, China
}

\begin{document}

\maketitle

\begin{abstract}

Physics-informed neural networks (PINNs)provide a powerful approach for solving partial differential equations (PDEs), but constructing a usable PINN remains labor-intensive and error-prone. Scientists must interpret problems as PDE formulations, design architectures and loss functions, and implement stable training pipelines. Existing large language model (LLM)approaches address isolated steps such as code generation or architecture suggestion, but typically assume a formal PDE is already specified and therefore lack an end-to-end perspective. We present \name, an LLM-driven multi-agent system that builds trainable PINNs directly from natural language task descriptions. \name\, coordinates four complementary agents: a \emph{PDE Agent} that parses task descriptions into symbolic PDEs, a \emph{PINN Agent} that selects architectures, a \emph{Code Agent} that generates modular implementations, and a \emph{Feedback Agent} that executes and diagnoses errors for iterative refinement. This design transforms informal task statements into executable and verifiable PINN code. Experiments show that \name\, achieves substantially lower errors and greater robustness than competitive baselines: mean squared error (MSE)is reduced by up to 3–5 orders of magnitude, end-to-end execution success improves by more than 50\%, and reduces time overhead by up to 74\%. 
\end{abstract}

\section{Introduction}
\label{section/1_introduction}

Partial differential equations (PDEs)are central to scientific computing, underpinning applications in physics, engineering, and materials science. Physics-informed neural networks (PINNs)\citep{raissi2019pinns} have emerged as a flexible framework that embeds governing equations into trainable neural models, offering a unified approach for forward, inverse, and data-scarce problems \citep{Karniadakis2021Physics,lu2019deepxde}. Despite their promise, training PINNs remains highly challenging: they suffer from gradient pathologies \citep{wang2021gradient}, ill-conditioning from the neural tangent kernel perspective \citep{wang2022ntk}, failure modes in complex regimes \citep{krishnapriyan2021characterizing}, and sensitivity to activation functions, sampling, and decomposition strategies \citep{jagtap2020adaptive,yu2021gpinn,wu2023rar,shukla2021xpinns,jagtap2020xpinns}. Although libraries and benchmarks such as DeepXDE \citep{lu2019deepxde}, PINNacle \citep{hao2023pinnacle}, and PDEBench \citep{takamoto2022pdebench} have been developed, deploying a trainable PINN still requires expert-level manual effort in PDE specification, architecture design, and optimization tuning.

Efforts to lower this barrier remain fragmented. Traditional automation focuses on hyperparameter search \citep{snoek2012bo,li2017hyperband,falkner2018bohb,HE2021AutoML} or architecture variants \citep{shukla2021xpinns,wang2023naspinn,wang2023autopinn}, but do not provide end-to-end workflows. Recent advances in large language models (LLMs)offer new possibilities: foundation models for code generation \citep{roziere2023codellama,li2023starcoder,lozhkov2024starcoder2} and agentic reasoning \citep{yao2023tree,shinn2023reflexion,madaan2023selfrefine,wang2023selfconsistency,wei2022CoT} enable natural-language interfaces to computational tasks. Domain-specific prototypes, such as CodePDE \citep{li2025codepde} and PINNsAgent \citep{wuwu2025pinnsagent}, demonstrate the feasibility of LLM-driven PDE solvers, but they often assume PDE schemas are given or lack verification and iterative refinement. Thus, a crucial gap remains: no existing system can start directly from natural language descriptions and deliver \emph{executable, verifiable, and trainable} PINN pipelines.

\begin{figure*}
    \centering
\includegraphics[width=\textwidth]{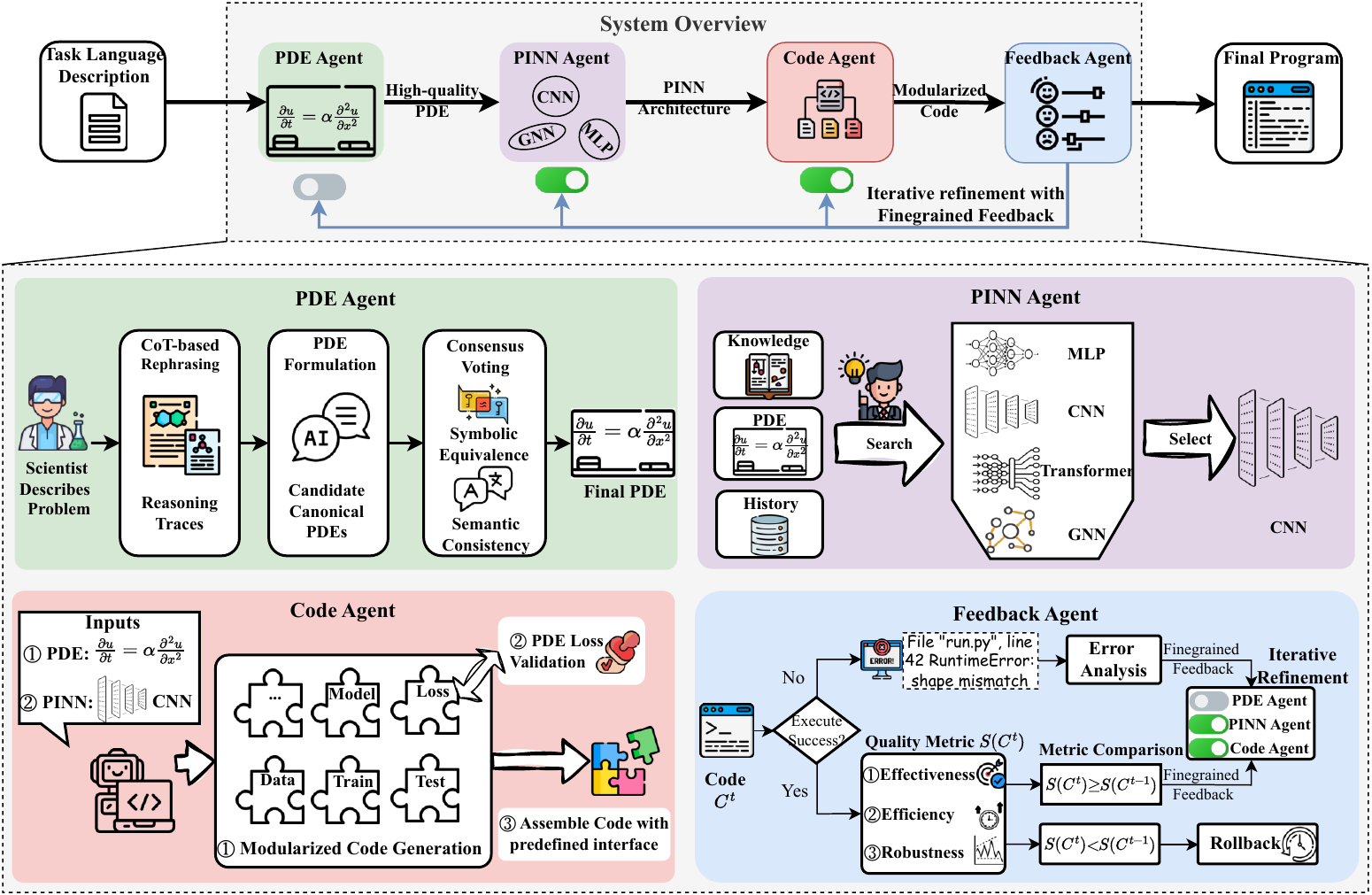}
    \caption{System overview of \textbf{\name}. 
The framework decomposes end-to-end PINN design into four agents: \pde\, (canonical PDE formulation), \pinn\, (training-free architecture selection), \code\, (modularized code generation), and \feedback\, (runtime error analysis and multi-dimensional evaluation). 
Iterative refinement with feedback forms a closed loop, yielding reliable and executable PINN programs from natural language descriptions.}
    \label{fig:system_overview}
    \vspace{-20pt}
\end{figure*}

To address this gap, we propose a multi-agent framework, namely \name, that decomposes the workflow into four cooperating roles, as shown in Fig.~\ref{fig:system_overview}. The \pde\, formulates natural language into operators, coefficients, and boundary/initial conditions. The \pinn\, aligns PDE characteristics—periodicity, geometric complexity, and multiscale or chaotic dynamics—with inductive biases via a requirement vector and utility score. The \code\, generates modular, contract-preserving training code, while the \feedback\, executes the code, monitors residuals and convergence, and iteratively guides corrections. This structured, verifiable pipeline ensures that scientific consistency, executability, and trainability are treated as first-class design goals.

Our contributions are as follows:
\begin{itemize}
    \item We propose the first framework that starts directly from natural language task descriptions and automatically produces complete PINN solutions, including PDE formulations, architecture selection, code generation, and feedback-driven refinement, thereby lowering the entry barrier for domain scientists.  
    \item We construct a benchmark dataset that pairs four-level difficulty task descriptions with ground-truth PDEs, enabling systematic evaluation of semantic-to-symbol grounding and supporting verifiable, reproducible PINN design.  
    \item We demonstrate that our multi-agent framework achieves substantial improvements across diverse PDEs, reducing mean squared error by up to \emph{3–5 orders of magnitude}, increasing code executability success rates by more than \emph{50\%}, and reducing time overhead up to \emph{74\%} compared to strong agent-based baselines.  
\end{itemize}

\section{Related Work}
\label{section/2_related_work}

\paragraph{Physics-Informed Neural Networks.}
Physics-Informed Neural Networks (PINNs)~\citep{raissi2019pinns} integrate governing equations into neural training by penalizing PDE residuals and boundary violations. Numerous variants improve convergence and accuracy through adaptive activations~\citep{jagtap2020adaptive}, gradient-enhanced residuals~\citep{yu2021gpinn}, adaptive sampling~\citep{lu2019deepxde,wu2023rar}, or domain decomposition~\citep{shukla2021xpinns,jagtap2020xpinns}. Yet, these approaches still require experts to manually specify PDE formulations, architectures, and loss terms. Our work instead seeks to automate these design choices from task descriptions.

\paragraph{LLM Agents and Reasoning Strategies.}
Large language and code models have enabled text-to-code generation~\citep{roziere2023codellama,li2023starcoder} and agentic software engineering~\citep{jimenez2023swebench,yang2024sweagent}. In scientific domains, CodePDE~\citep{li2025codepde} demonstrates that inference-time reasoning and self-debugging can produce PDE solvers directly from text. Complementary prompting strategies such as SCoT~\citep{li2025structured} and Self-Debug~\citep{chen2023teaching} improve logical consistency and error correction through structured reasoning or iterative reflection. However, these remain single-agent methods without physics-grounded validation, limiting their applicability to scientific surrogates. Our framework extends this direction by coupling reasoning and feedback across multiple specialized agents tailored to PINNs.

\paragraph{Automated PINN Design.}
Classical Automated Machine Learning (AutoML)methods~\citep{HE2021AutoML}, including Bayesian optimization~\citep{snoek2012bo}, Hyperband~\citep{li2017hyperband}, and BOHB~\citep{falkner2018bohb}, aim to reduce manual effort in tuning architectures and hyperparameters. Applied to physics-informed settings, however, they struggle with residual imbalance, unit inconsistency, and multi-scale stiffness, often requiring expert intervention. Recent PINN-oriented searches~\citep{wang2021gradient,wu2023rar} mitigate some challenges but still assume human-specified PDEs and loss structures. In contrast, our approach introduces a dedicated multi-agent system for PINN automation, integrating PDE translation, architecture design, and feedback-driven refinement to minimize manual design effort and achieve end-to-end trainability.


\section{Motivation}
\label{sec:causal_analysis}

Despite recent progress on PDE parsing and PINN architecture search, robust end-to-end automation remains elusive. Existing studies tend to optimize individual steps in isolation, overlooking systematic dependencies across the pipeline. Our empirical analysis reveals three fundamental bottlenecks that persist in practice, each supported by controlled experiments. We describe these challenges below to motivate the design choices introduced in the Sec.~\ref{sec:method}.

\subsection{Ambiguity of Translating Tasks into PDEs}\label{sec:motivation_PDE}
 
\begin{wrapfigure}{r}{0.5\textwidth} 
    \centering
    \vspace{-10pt} 
    \includegraphics[width=0.5\textwidth]{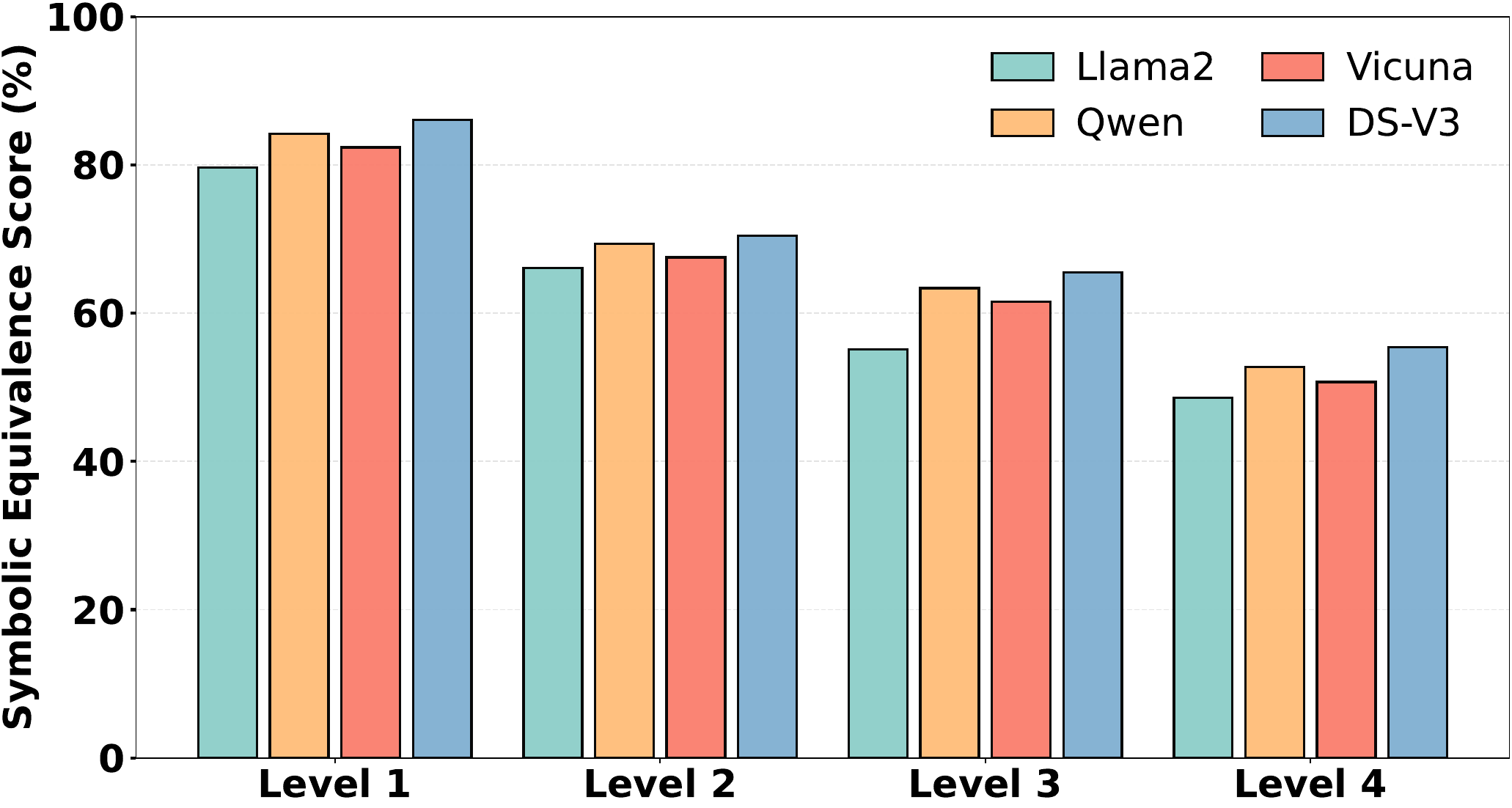}
    \caption{
    Impact of linguistic complexity on PDE translation. Accuracy is reported across four levels of description difficulty using Symbolic Equivalence and Semantic Consistency scores.
    }
    \vspace{-10pt}
    \label{fig:diff}
\end{wrapfigure}

The pipeline begins with translating a natural-language description into a formal PDE, which defines the loss terms, constrains the solution space, and conditions all downstream stages. Any error in this step renders the pipeline invalid, making reliable formulation essential.

To examine this challenge, we construct the \textbf{Task2PDE} dataset by selecting eight PDEs from PINNacle benchamrk~\citep{hao2023pinnacle} and re-expressing each at four levels of linguistic variability (Level~1 = simplest, Level~4 = most complex; see Appendix~\ref{app:task2pde}). Each PDE is paired with 50 descriptions per level, yielding 1,600 description–equation pairs. Four open-source LLMs (Llama2~\citep{touvron2023llama2openfoundation}, Vicuna~\citep{vicuna2023}, DeepSeek-V3~\citep{deepseekai2025deepseekv3}, Qwen~\citep{bai2023qwen1})are evaluated using \emph{symbolic equivalence} (introduced in Sec.~\ref{sec:PDE-agent}). Results in Fig.~\ref{fig:diff} show that symbolic accuracy consistently degrades as descriptions grow more complex, reflecting the fragility of direct PDE formulation from natural language.

However, symbolic equivalence is overly strict: mathematically identical equations, such as $u_{xx}$ and $\partial^2 u/\partial x^2$, are judged inconsistent, and alternative coefficient expressions are misclassified. This underestimates true capability and prevents reliable filtering under noisy inputs, undermining downstream PINN design and code generation. These observations suggest that symbolic checks alone are insufficient, motivating the \emph{PDE Agent} in Sec.~\ref{sec:PDE-agent}, which integrates semantic evaluation and consensus voting to ensure robust PDE formulation.

\vspace{-5pt}

\subsection{Variability of Architecture Performance across PDEs}\label{sec:motivation_PINN}

\begin{wrapfigure}{r}{0.5\textwidth}
    \centering
\includegraphics[width=0.5\textwidth]{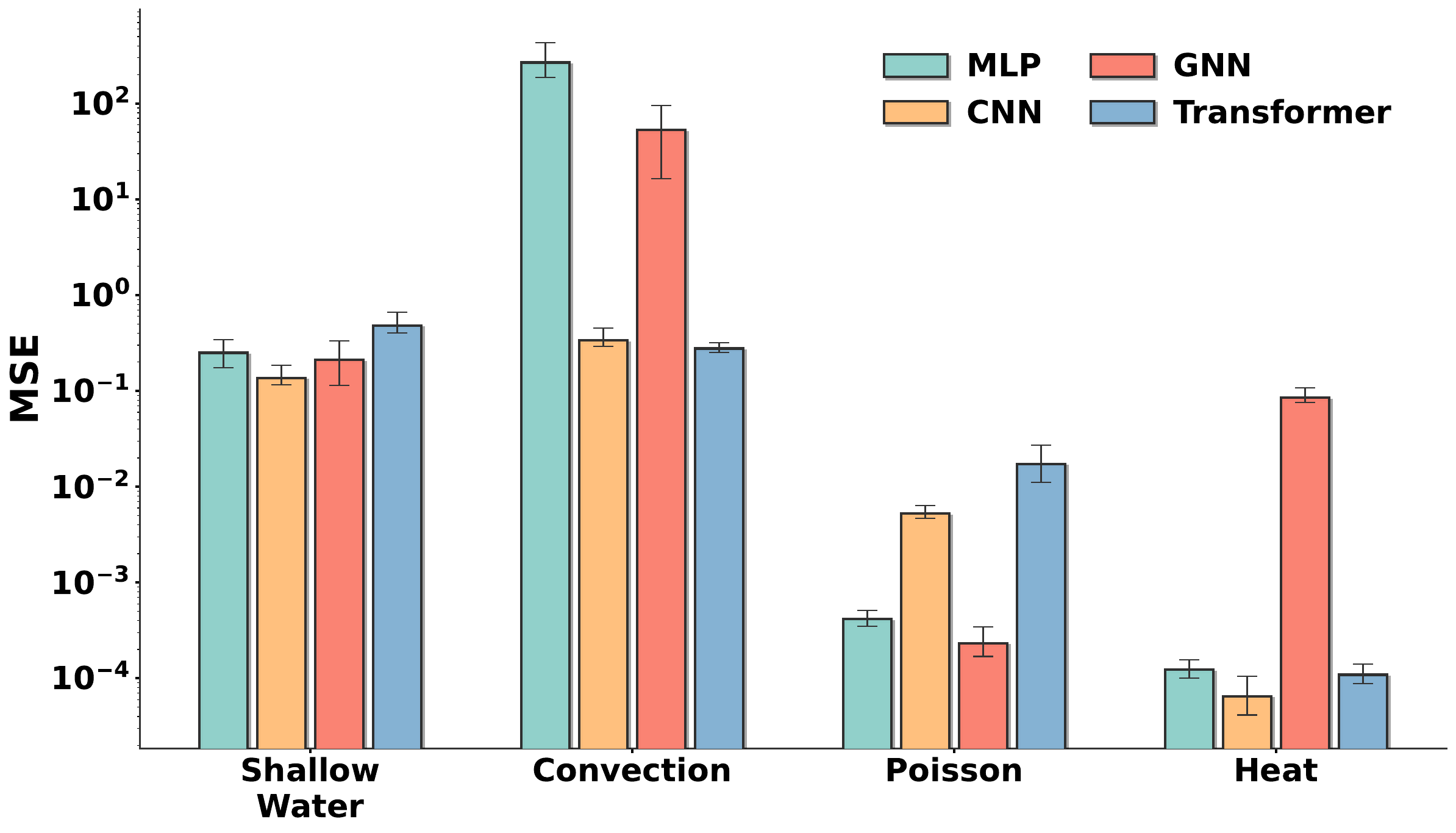}
   
    \vspace{-5pt}
    \caption{Comparative MSE of different PINN architectures on representative PDEs.
    Results are shown in log scale for clarity.}
    \vspace{-10pt}
    \label{fig:ab_all}
\end{wrapfigure}

Once the PDE is specified, selecting a suitable PINN architecture is crucial. The inductive bias of the network, such as its preference for local patterns, long-range dependencies, or structural constraints, directly affects stability and accuracy. A poor match can lead to slow convergence or large residual errors. To demonstrate this effect, we benchmark four representative architectures (MLP, CNN, GNN, and Transformer)on PDEs including Shallow Water, Convection, Poisson, and Heat. As shown in Fig.~\ref{fig:ab_all}, performance varies markedly across PDEs. CNNs and Transformers excel on Convection and Heat, while MLPs and GNNs achieve the lowest error on Poisson. For Shallow Water, differences are minor. These results show that no single architecture is universally effective, motivating approaches that adapt PINN designs to the operators and structures of different PDEs.

\begin{wrapfigure}{r}{0.45\textwidth}
    \centering
\vspace{-10pt}    
\includegraphics[width=0.45\textwidth]{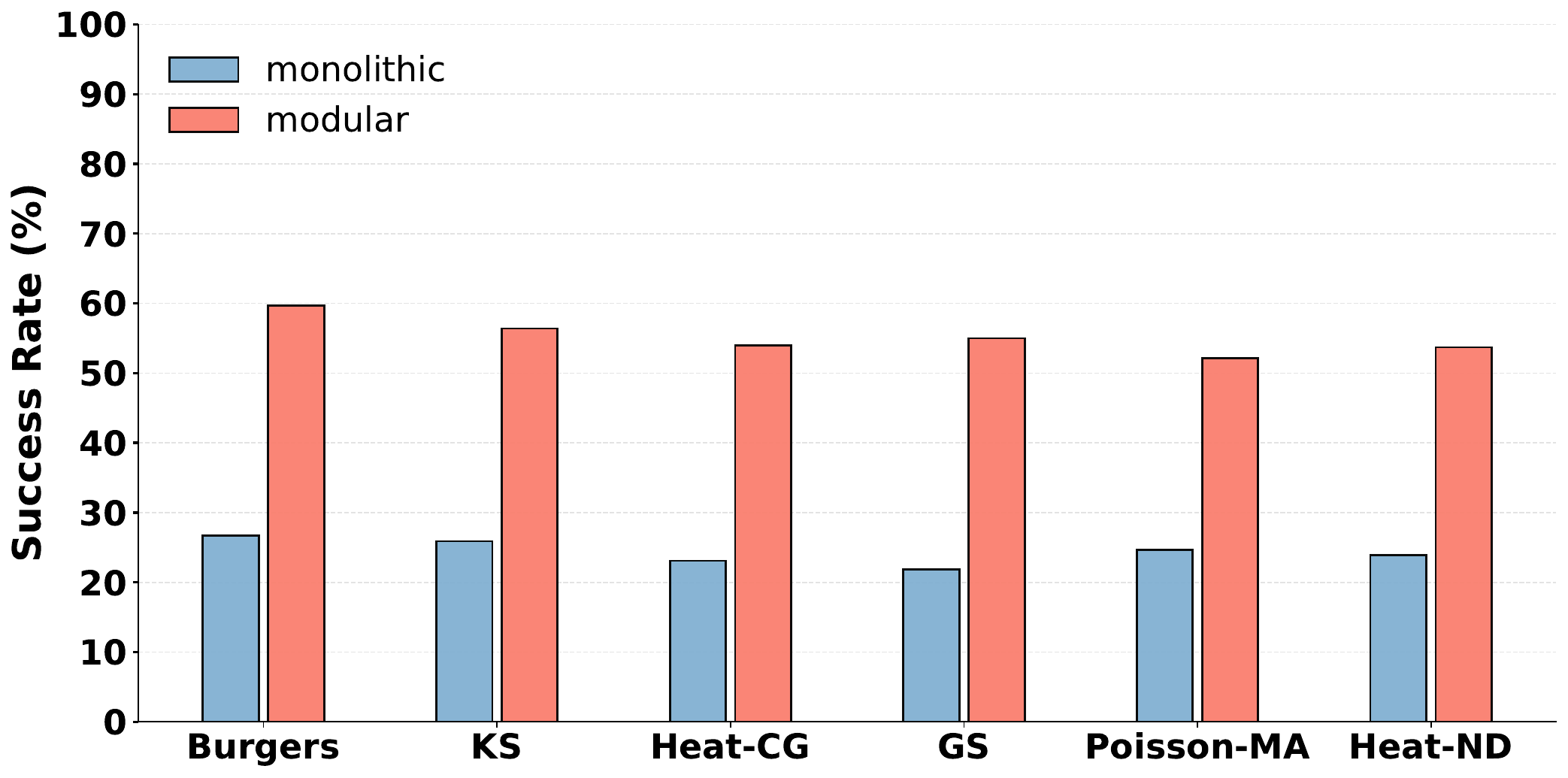}
   
    \caption{Comparative Success Rate(\%) of different code generation paradigms (monolithic vs. modular)on six  PDEs.
    }
    \vspace{-5pt}
    \label{fig:code}
\end{wrapfigure}
\subsection{Complexity of Code Generation in End-to-End Workflows}\label{sec:motivation_code}



After the PDE and PINN architecture are specified, the next step is to generate executable code, including model definitions, physics-informed losses, data pipelines, and training routines. This process is complex because multiple components must not only be correct in isolation but also interact reliably, making executability a central challenge.

To study code generation paradigms, we compare \emph{monolithic generation}, where an LLM produces the entire pipeline in a single pass, with \emph{modular generation}, where code is synthesized by components. As shown in Fig.~\ref{fig:code}, modular generation consistently achieves more than twice the success rate of monolithic generation across six representative PDEs (Burgers, KS, Heat-CG, GS, Poisson-MA, Heat-ND). The modular design localizes errors, preserves correct components, and avoids regenerating the full script, thereby substantially improving executability. These results motivate the design of the \textbf{Code Agent}, which adopts the modular paradigm. We note that this experiment isolates the effect of modularization alone; when combined with the \textbf{Feedback Agent} in our full framework, success rates improve even further, as shown in later sections.

\section{Method}\label{sec:method}

\subsection{System Overview}

Fig.~\ref{fig:system_overview} presents \name, our multi-agent framework that converts natural-language task descriptions into executable PINN training code. It consists of four agents with distinct roles: the \emph{PDE Agent} formalizes task descriptions into governing equations, the \emph{PINN Agent} selects suitable architectures, the \emph{Code Agent} generates modular implementations, and the \emph{Feedback Agent} executes and evaluates outputs.   These agents interact in a sequential workflow, with the \feedback\, providing iterative diagnostics that refine earlier stages, particularly code generation. This modular and feedback-driven design reduces error propagation and ensures reliable, scientifically valid PINN implementations.



\subsection{PDE Agent}
\label{sec:PDE-agent}

To alleviate the sensitivity to linguistic variability identified in Sec.~\ref{sec:motivation_PDE}, the \emph{PDE Agent} uses a label-free reasoning–selection pipeline. Given a task description $d$, the agent samples $K$ chain-of-thought (CoT)trajectories, cleans each trajectory into a normalized description $\hat d_k$, and formulates a canonical PDE candidate $E_k$. Invalid candidates are filtered by template validation (operator well-formedness, residual form, admissible boundary/initial terms). The remaining set $\mathcal{E}=\{E_1,\dots,E_M\}$ is then resolved via consensus voting, and the agent selects the candidate that is most similar to the others under a joint symbolic–semantic criterion.

\textbf{Symbolic Equivalence.}
To assess whether two candidate PDEs express the same operator structure, we compute a \emph{symbolic equivalence score} based on their abstract syntax trees (ASTs). Each PDE $E$ is parsed into a canonical symbolic tree $\mathcal{T}(E)$ using Sympy, where nodes represent operators (e.g., $\partial_t$, $\partial_x^2$, nonlinear products)and leaves correspond to variables or constants. 

Given two trees $\mathcal{T}(E_i)$ and $\mathcal{T}(E_j)$, we define their symbolic equivalence as a normalized tree-matching score,
\begin{equation}
\mathrm{sym}(E_i,E_j)
=\frac{|\mathcal{M}(\mathcal{T}(E_i),\mathcal{T}(E_j))|}
{\max\!\big(|\mathcal{T}(E_i)|,\,|\mathcal{T}(E_j)|\big)},
\end{equation}
where $\mathcal{M}(\mathcal{T}(E_i),\mathcal{T}(E_j))$ denotes the set of matched subtrees under operator-preserving alignment, and $|\mathcal{T}(\cdot)|$ counts the total nodes. This yields a score in $[0,1]$, equal to $1$ if two PDEs are symbolically equivalent (identical operator trees)and decreasing smoothly as structural discrepancies grow. 

This formulation abstracts our Sympy-based implementation, where equivalence is resolved by recursively comparing operator nodes and their children up to commutativity and normalization rules. It aligns with symbolic regression principles~\citep{rudy2017data,la2021contemporary}, while providing robustness to variations in coefficient presentation or term ordering.


\textbf{Semantic Consistency.} Symbolic matching alone cannot capture cases where mathematically equivalent PDEs are expressed in different notations or variable names. Following ideas from mathematical information retrieval~\citep{zanibbi2016multi}, we therefore introduce a \emph{semantic consistency} score. Each candidate PDE $E$ is paraphrased into a normalized summary $g(E)$ that encodes its domain, operator types, and forcing terms.  The semantic consistency between two candidates $E_i$ and $E_j$ is then defined as
\begin{equation}
\mathrm{sem}(E_i,E_j)=\sigma\!\big(g(E_i),\,g(E_j)\big),
\end{equation}

\noindent where $\sigma$ is a sentence-level similarity function such as embedding cosine similarity or LLM-based entailment scoring. This yields values in $[0,1]$ and provides robustness to symbol renaming, coefficient scaling, or algebraic rearrangements that preserve meaning but alter surface form.

\textbf{Consensus Voting.}
Finally, we combine symbolic and semantic similarities into a composite score 
$S(E_i,E_j)=\alpha\,\mathrm{sym}(E_i,E_j)+(1-\alpha)\,\mathrm{sem}(E_i,E_j)$. 
Each candidate is then compared against the others, and the one with the highest average similarity is selected as the final PDE. 
This simple consensus step ensures that the chosen equation is both structurally consistent and semantically faithful to the task description.


\subsection{PINN Agent}\label{sec:PINN-Agent}

Different PDEs exhibit distinct sensitivities to network architecture, with no single structure performing best across tasks, as shown in Sec.~\ref{sec:motivation_PINN}. The \emph{PINN Agent} addresses this challenge by framing architecture selection as a \emph{training-free, test-time reasoning problem}. Its inputs include the canonical PDE representation $E$ from the PDE Agent, a knowledge base $\mathcal{K}$ encoding heuristic priors, and a history cache $\mathcal{H}$ recording past architecture–performance outcomes.

\paragraph{History reuse.}

When a new PDE $E$ is encountered, the agent first queries $\mathcal{H}$ to determine whether a similar equation has been previously solved. Similarity is judged through LLM reasoning over semantic forms and boundary conditions. If a match exists, the previously selected architecture $\mathcal{A}$ is reused directly, avoiding redundant search. If no suitable match is found, the agent falls back to a more general strategy, namely the \emph{knowledge-guided matching} introduced next, which derives architecture choices from the physical characteristics of the PDE itself.

\paragraph{Knowledge-guided Matching.}
In the absence of reusable history, the agent selects architectures via \emph{knowledge-guided matching}. The key idea is to embed PDEs and architectures into a shared representation space, where their alignment can be systematically evaluated. We first describe how PDEs are represented, then how architectures are encoded, and finally how the two are matched.

\textit{1. PDE Feature Representation.}  To represent the input side of the matching process, each PDE $E$ is encoded as a feature vector
\begin{equation}
\phi(E)= [f_1(E), f_2(E), \dots, f_n(E)]^\top,
\end{equation}
where $f_i(E)$ denotes a quantifiable physical property. Representative examples include \emph{periodicity}, \emph{geometry complexity}, and \emph{multi-scale demand}. Periodicity reflects whether domains or boundary conditions repeat, geometry complexity captures whether the domain is structured or irregular, and multi-scale demand indicates the extent of interacting scales or chaotic regimes. Formal definitions are given in Appendix~\ref{app:pinnagent-metrics}. These dimensions are motivated by prior findings that Fourier or sinusoidal layers align with periodic problems~\citep{sitzmann2020implicit,li2020fourier}, graph-based models are effective for irregular geometries~\citep{pfaff2020learning,brandstetter2022message}, and attention or spectral operators handle multi-scale dynamics~\citep{rahman2024pretraining}.  

\textit{2. Architecture Capability Representation.}  
To make architectures comparable with PDE features, each architecture $\mathcal{A}$ is represented by a capability vector
\begin{equation}
\psi(\mathcal{A})= [a_1(\mathcal{A}), a_2(\mathcal{A}), \dots, a_n(\mathcal{A})]^\top,
\end{equation}
where $a_i(\mathcal{A})$ measures its competence on property $i$. Capability values are inferred through LLM reasoning and refined with historical experimental outcomes, ensuring adaptability across tasks.

\textit{3. PDE–Architecture Matching}  The compatibility between a PDE $E$ and an architecture $\mathcal{A}$ is measured using a weighted cosine similarity:
\begin{equation}
S(\mathcal{A}, E)=
\frac{(\mathbf{W}\phi(E))^\top \psi(\mathcal{A})}
{\|\mathbf{W}\phi(E)\|_2 \cdot \|\psi(\mathcal{A})\|_2},
\end{equation}
where $\mathbf{W}=\mathrm{diag}(w_{\mathrm{per}},w_{\mathrm{geo}},w_{\mathrm{ms}})$ assigns importance weights to each property. In practice, we prioritize multi-scale demand over geometry and periodicity, as mismatches on the former are most detrimental to convergence~\citep{li2020fourier, pfaff2020learning, brandstetter2022message, sitzmann2020implicit}. The final architecture is then selected as
\begin{equation}
\mathcal{A}^\star = \arg\max_{\mathcal{A} \in \Theta} S(\mathcal{A} \mid E).
\end{equation}

\subsection{Code Agent }
\label{sec:Code-Agent}

Directly prompting an LLM to generate the entire PINN pipeline in one pass often produces brittle code, where model definition, loss formulation, and training loops are tightly coupled. Errors become difficult to isolate, and fixing them typically requires regenerating the whole script. To avoid this, the \emph{Code Agent} adopts a modular strategy with explicit verification mechanisms. 

\textbf{Modularized code generation.}
Instead of producing a monolithic script, the \code\, decomposes the pipeline into independent modules: (i)model definition, (ii)PDE loss, (iii)data preprocessing, (iv)training loop, (v)validation, and (vi)main function. Each module is generated separately, allowing faults to be localized and corrected without regenerating unrelated components.

\textbf{Interface constraints.}
Modules are connected through standardized input–output formats, ensuring compatibility and composability. This design makes it possible to update or replace one module without introducing inconsistencies elsewhere, thereby reducing correction cost and enabling fine-grained refinement.

\textbf{PDE loss verification.}
For the PDE loss module, the generated code is parsed back into a symbolic PDE $\hat{E}$ and checked for equivalence with the PDE $E$ provided by the \pde. Only loss modules that pass this symbolic check are retained, ensuring that the optimization objective faithfully encodes the governing equation.

\subsection{Feedback Agent}
\label{sec:Feedback-Agent}

The \emph{Feedback Agent} closes the loop by leveraging runtime signals to refine earlier stages. Built on the modular code of the \code, it translates execution diagnostics into localized suggestions, avoiding global regeneration and improving reliability.


\textbf{Error localization and correction.}
When executing the generated code, two scenarios arise. If runtime errors occur, the \feedback\, analyzes the error messages and attributes them to the most likely module (e.g., model structure, loss function, training loop). It then instructs the \code\, to regenerate only the faulty component, avoiding unnecessary changes to other modules. If the issue originates upstream (e.g., in PDE specification or PINN architecture), the \feedback\, can escalate its directive to the corresponding agent, ensuring that corrections are applied at the appropriate level.

\textbf{Multi-dimensional quality evaluation.}
If execution succeeds, the \feedback\, evaluates the code along three complementary dimensions: (i)\emph{effectiveness}, measured by PDE residual error (e.g., MSE); (ii)\emph{efficiency}, measured by convergence speed and resource cost (steps, FLOPs, parameters); and (iii)\emph{robustness}, measured by loss smoothness and the absence of gradient pathologies. Each metric is normalized, and a weighted sum produces an overall quality score:
\begin{equation}
S(C)= \sum_{i=1}^{3} w_i \, \hat{m}_i(C),
\end{equation}
where $C$ denotes the generated code, $\hat{m}_i$ the normalized value of the $i$-th metric, and $w_i$ its weight. Detailed definitions and quantification of these metrics are provided in Appendix~\ref{app:feedback-metrics}.

\textbf{Iterative refinement.}
The decision to accept or reject a new version is based on comparing the current score $S(C^{(t)})$ with the previous score $S(C^{(t-1)})$. If the new version improves, the agent proceeds; otherwise, it reverts and restarts optimization. By coupling modular generation with runtime feedback, the system ensures that diagnostic signals can be acted upon locally rather than globally, providing fine-grained corrections that improve reliability and efficiency over iterations.

\section{Experiments}

\subsection{Experimetal Settings}

\paragraph{Benchmark Datasets}

We evaluate \name\, on the PINNacle benchmark~\citep{hao2023pinnacle}, which comprises 14 representative PDEs across 1D, 2D, 3D, and ND settings: \emph{Burgers, Wave-C, KS, Burgers-C, Wave-CG, Heat-CG, NS-C, GS, Heat-MS, Heat-VC, Poisson-MA, Poisson-CG, Poisson-ND, Heat-ND}. This collection spans diverse dimensionalities, geometric complexities, and dynamical regimes, providing a rigorous testbed for automated PINN design. At the task-to-PDE stage, \name\, operates from natural-language inputs: for each PDE we construct three distinct textual problem descriptions, which must be translated into canonical PDE formulations before downstream modeling. In contrast, baseline methods cannot perform this translation step and are therefore provided directly with the canonical PDE formulations from the benchmark. Each task is evaluated over 10 independent runs, and within each run the agent is allowed up to three refinement iterations, ensuring both fairness across methods and robustness to stochasticity in generation.

\begin{wraptable}{r}{0.45\textwidth}
\vspace{-22pt}
\centering
\caption{Comparison of methods across five functional dimensions: 
\textbf{PF} (PDE formulation), 
\textbf{AD} (architecture design),  
\textbf{CG} (code generation), 
and \textbf{FS} (feedback signal). For feedback signal, ``Err+Metrics'' augments runtime error with validation metrics.
}

\label{tab:bstype}
\small
\renewcommand{\arraystretch}{1.2}
\setlength{\tabcolsep}{6pt}
\scalebox{0.88}{

\begin{tabular}{lcccc}
\toprule
\textbf{Method} & \textbf{PF} & \textbf{AD}  & \textbf{CG} & \textbf{FS} \\
\midrule
PINNacle      & \xmark  & \xmark & \xmark                & \xmark \\
RandomAgent   & \xmark & \cmark & Partial& \xmark \\
BayesianAgent & \xmark  & \cmark & Partial& \xmark \\
SCoT          & \xmark  & \xmark & Partial           & \xmark \\
Self-Debug    & \xmark  & \xmark & Partial           & Err-only \\
PINNsAgent    & \xmark  & \cmark & Full              & Err+Metrics \\
\hline
\textbf{Lang-PINN}     & \cmark & \cmark & Full              & Err+Metrics \\
\bottomrule
\end{tabular}
}
\end{wraptable}

\paragraph{Baselins}

We include \textbf{PINNacle}~\citep{hao2023pinnacle} as a non-agent reference that fixes both PDEs and architectures and directly trains PINNs. All other baselines adopt LLM-based agent but still assume the PDE and architecture are given. \textbf{RandomAgent} and \textbf{BayesianAgent} explore architectures  through random or Bayesian search with error-only feedback, while \textbf{SCoT}~\citep{li2025structured}, \textbf{Self-Debug}~\citep{chen2023teaching}, and \textbf{PINNsAgent}~\citep{wuwu2025pinnsagent} rely on prompting to generate losses or partial code, again without full feedback or PDE formulation.  As summarized in Table~\ref{tab:bstype}, none of these baselines support PDE formulation, code generation is at best partial, and feedback is limited to error detection, whereas \textbf{Lang-PINN} spans all dimensions in a coordinated multi-agent system.

\subsection{Main Results}



\renewcommand{\arraystretch}{1.22}   
\setlength{\tabcolsep}{1.6pt}        
\small

\begin{table*}[!ht]
\centering
\caption{Comparative performance (MSE)on 14 different PDEs (averaged over 10 runs).}
\label{tab:main_results_fixed}
\scalebox{0.85}
{
\begin{tabular}{clcccccc|c}
\hline
\textbf{Dim} & \textbf{PDE} & \textbf{RandomAgent} & \textbf{BayesianAgent} & \textbf{PINNsAgent}  & \textbf{SCoT} & \textbf{Self-Debug} & \textbf{Ours} & \textbf{PINNacle}\\
\hline
\multirow{3}{*}{\textbf{1D}}
  & Burgers   & 6.63E-02 & 8.70E-02 & 1.10E-04  & 1.40E+01 & 1.26E+01 & \textbf{6.48E-05} & \textcolor{gray}{7.90E-05} \\ 
  & Wave-C    & 1.50E-01 & 1.78E-01 & 3.74E-02  & 1.28E+00 & 1.18E+00 & \textbf{2.25E-03} & \textcolor{gray}{3.01E-03}\\ 
  & KS        & 1.09E+00 & 1.10E+00 & 1.09E+00 & 3.33E+00 & 2.93E+00 & \textbf{1.62E-03} & \textcolor{gray}{1.04E+00} \\
\hline
\multirow{8}{*}{\textbf{2D}}
  & Burgers-C & 2.48E-01 & 2.42E-01 & 2.93E-01  & 4.54E-01 & 4.09E-02 & \textbf{2.88E-03} & \textcolor{gray}{1.09E-01}\\ 
  & Wave-CG   & 2.87E-02 & 2.11E-02 & 4.59E-02  & 2.00E+00 & 1.90E+00 & \textbf{2.52E-03} & \textcolor{gray}{2.99E-02}\\ 
  & Heat-CG   & 3.96E-01 & 1.17E-01 & 9.06E-02  & 4.38E+00 & 3.81E-02 & \textbf{1.35E-03} & \textcolor{gray}{8.53E-04}\\ 
  & NS-C      & 4.02E-03 & 5.12E-03 & \textbf{1.40E-05} & 5.67E-01 & 5.27E-01 & 4.05E-05 & \textcolor{gray}{2.33E-05} \\ 
  & GS        & 4.28E-03 & 4.03E-03 & 3.37E+08 & 3.76E+00 & 3.35E+00 & \textbf{1.89E-03} & \textcolor{gray}{4.32E-03} \\ 
  & Heat-MS   & 1.84E-02 & 7.48E-03 & 1.06E-04 & 7.10E-02 & 6.04E-03 & \textbf{2.27E-05} & \textcolor{gray}{5.27E-05} \\ 
  & Heat-VC   & 3.57E-02 & 3.93E-02 & 1.43E-02  & 4.46E+00 & 4.01E-02 & \textbf{1.62E-03} & \textcolor{gray}{1.76E-03}\\ 
  & Poisson-MA& 5.87E+00 & 5.82E+00 & 3.16E+00 & 1.24E+04 & 1.07E+04 & \textbf{2.25E-03} & \textcolor{gray}{1.83E+00} \\
\hline
\multirow{1}{*}{\textbf{3D}}
  & Poisson-CG& 3.82E-02 & 2.55E-02 & 3.35E-02  & 4.17E-02 & 9.51E-03 & \textbf{1.35E-03} & \textcolor{gray}{9.51E-04}\\
\hline
\multirow{2}{*}{\textbf{ND}}
  & Poisson-ND& 1.30E-04 & 4.72E-05 & 4.77E-04 & 9.93E+00 & 9.43E+00 &  \textbf{8.42E-06}  & \textcolor{gray}{2.09E-06}\\ 
  & Heat-ND   & 2.58E-00 &  \textbf{1.18E-04} & 8.57E-04 & 3.74E+00 & 3.40E-03 & 4.72E-04 & \textcolor{gray}{8.52E+00} \\
\hline
\end{tabular}
}
\vspace{-15pt}
\end{table*}


\textbf{MSE Results.}
Table~\ref{tab:main_results_fixed} shows that \textbf{Lang-PINN} achieves the lowest errors on most PDEs, despite being the only approach that must first infer PDE formulations from natural language descriptions. 
In contrast, \emph{PINNacle} represents a human-expert–designed reference, where both the governing PDEs and PINN architectures are fixed in advance. 
Even against this strong baseline, \name delivers significant improvements. For instance, errors on \emph{KS} (1D), \emph{Poisson-MA} (2D), and \emph{Heat-ND} (ND)are reduced by over three orders of magnitude. 
Compared to agent-based baselines, the advantage is equally clear: while their errors on \emph{KS} and \emph{Poisson-MA} remain around $10^0$ to $10^4$, Lang-PINN reaches $10^{-3}$, demonstrating far stronger fidelity in solution quality.


\begin{wrapfigure}[12]{r}{0.48\textwidth} 
  \vspace{-4pt}                          
  \centering
  \includegraphics[width=\linewidth]{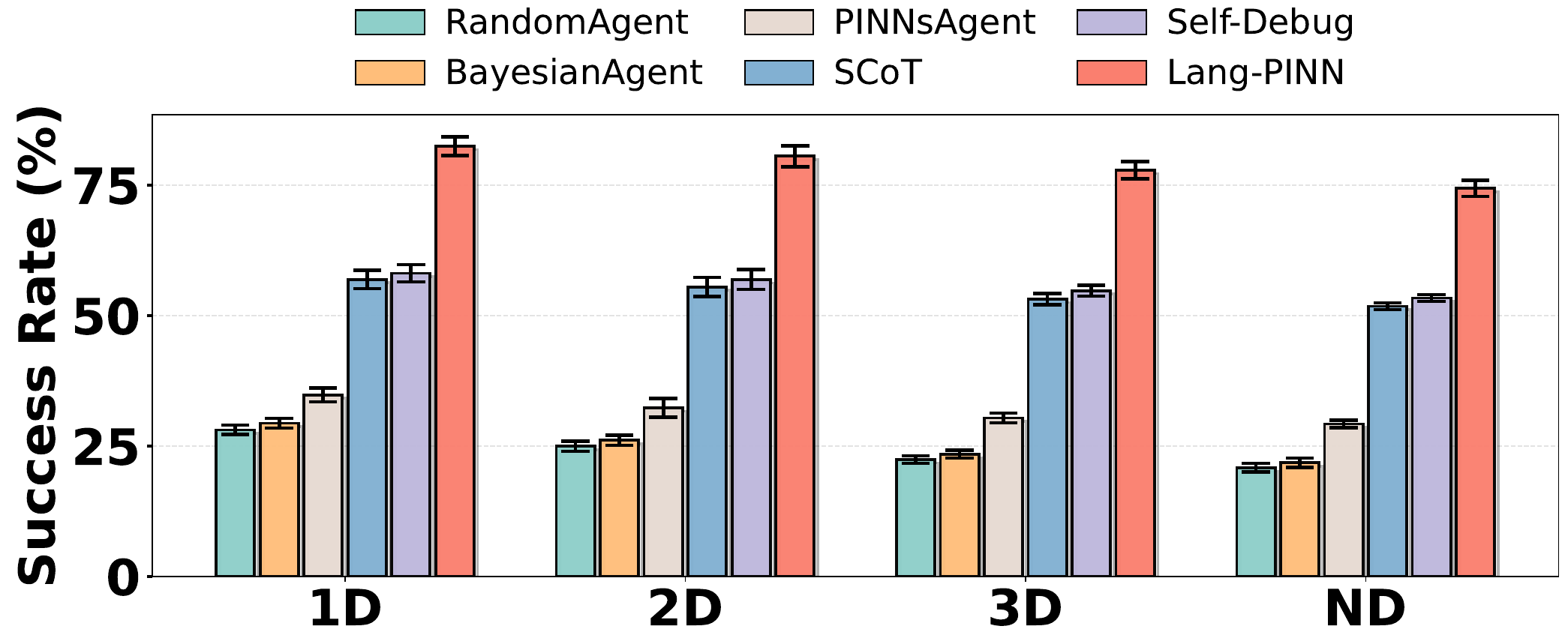}
  \caption{Comparative success rates (\%)across 1D/2D/3D/ND.}
  \label{fig:all_success}
  \vspace{-4pt}                          
\end{wrapfigure}

\enlargethispage{1.2\baselineskip}

\textbf{Success Rate.} Fig.~\ref{fig:all_success} reports the average success rate across PDEs of different dimensionalities. 
\name\, consistently delivers the highest reliability, with success exceeding 80\% in 1D and 2D regimes where baselines such as RandomAgent, BayesianAgent, and PINNsAgent typically remain below 35\%. 
Performance also remains robust in 3D, where \name\, maintains success rates close to 75\%, much higher than all baselines. 

\textbf{Time Overhead.} We evaluate efficiency by measuring the number of iterations required to obtain executable PINNs, with all methods capped at 50 iterations for fairness. Our \name\ converges in only 8 iterations on average, which is about 74\% fewer than the worst baseline (31), demonstrating substantial efficiency gains. Compared to other methods such as BayesianAgent (29), PINNsAgent (21), SCoT (17), and Self-Debug (14), our \name\, consistently reduces iteration counts, confirming that the joint design of modular code generation and feedback refinement accelerates convergence across diverse PDEs.


\subsection{Ablation Studies}


\paragraph{The Impact of PDE Agent}

\begin{figure}[!htb]
  \centering
  \begin{minipage}{0.4\textwidth}
    \centering
\includegraphics[width=\linewidth]{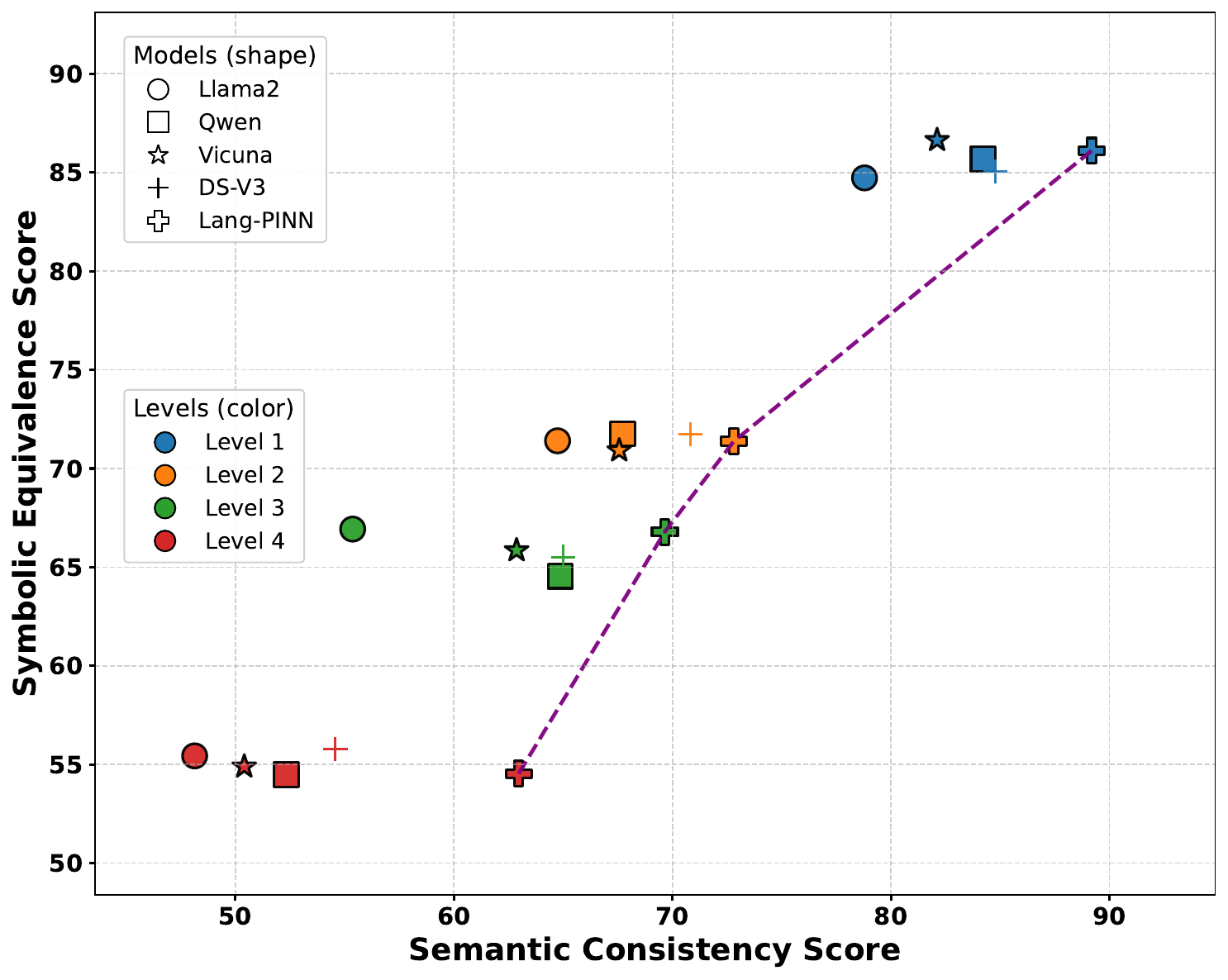}
      \vspace{-10pt}
    \caption{PDE formulation accuracy under four levels of linguistic complexity among different LLMs.  Our method (\name)lies on the Pareto frontier, achieving balanced improvements in both symbolic equivalence and semantic consistency scores.
    }
    \label{fig:ab_diff}
  \end{minipage}\hfill
  \begin{minipage}{0.58\textwidth}
    \centering
    
\includegraphics[width=\textwidth]{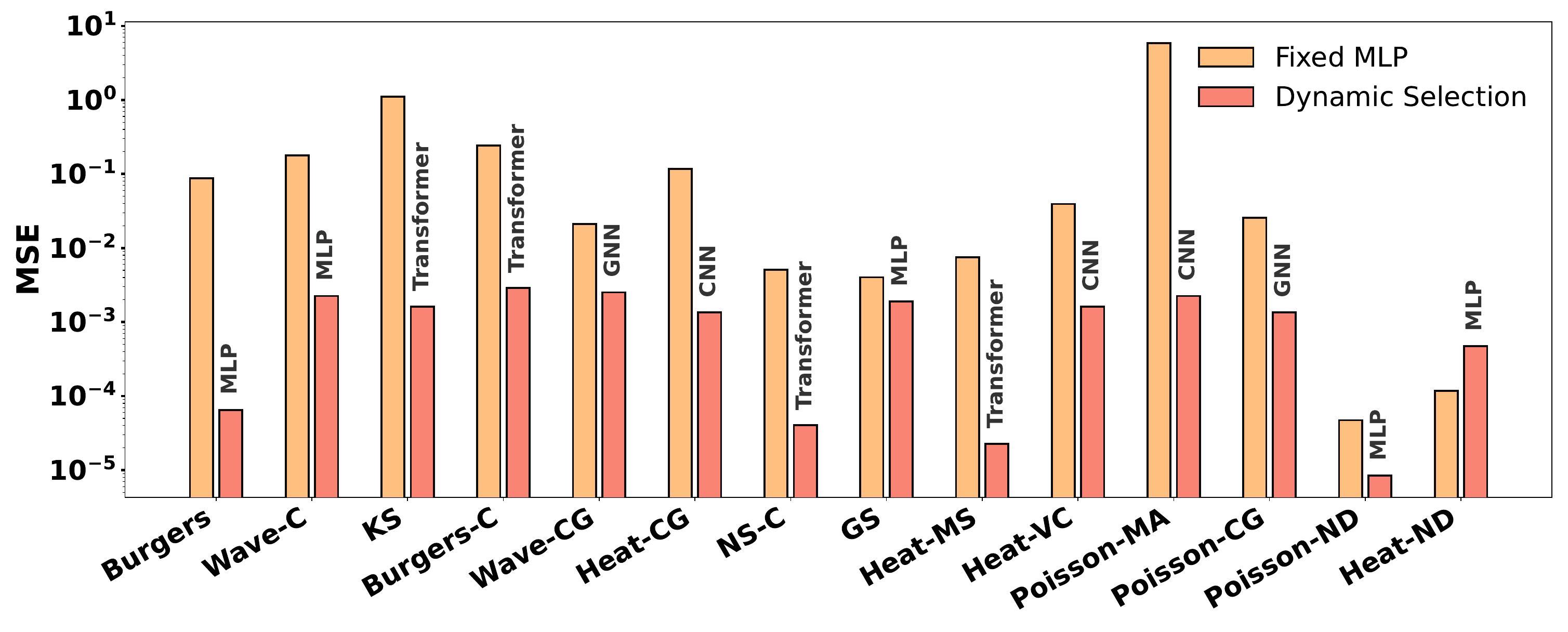}
   
    \caption{Ablation study of the PINN Agent: the fixed MLP variant (yellow)uses the same MLP backbone for all PDEs while the full PINN Agent (red)dynamically selects among diverse candidate architectures (e.g., MLP, CNN, and GNN). Dynamic selection consistently reduces MSE across 14 PDEs, demonstrating the effectiveness of adaptive architecture design.
    }
    \label{fig:ab_pinn}
  \end{minipage}
\end{figure}

Since Sec.~\ref{sec:motivation_PDE} highlighted the difficulty of faithfully grounding natural-language descriptions into PDEs, we conduct an ablation study to assess the contribution of our proposed PDE Agent. 
Fig.~\ref{fig:ab_diff} illustrates translation accuracy under increasing linguistic complexity. While all baselines degrade sharply from Level~1 to Level~4, our full agent consistently achieves the highest semantic consistency and maintains competitive symbolic equivalence. The gains are most evident under noisy and fragmented settings, where reasoning–canonicalization–validation steps prevent collapse and self-consistency selection stabilizes outputs. This demonstrates that the \pde\, not only alleviates sensitivity to surface-form variation but also provides robust task-to-equation translation, complementing the improvements observed in MSE and executable success rate.

\vspace{-10pt}

\paragraph{The Impact of PINN Agent.} 
To evaluate the contribution of the \pinn\, in dynamically selecting architectures, we compare it with a variant where the architecture is fixed to an MLP across all PDEs, with only depth and width tuned. In contrast, the \pinn\, leverages PDE, prior knowledge, and history to select among different architecture families (MLP, CNN, GNN, and Transformer). As shown in Fig.~\ref{fig:ab_pinn}, dynamic selection achieves substantially lower MSEs across 14 PDEs, with the largest gains on periodic, irregular, or multi-scale problems (KS, Poisson-MA, Heat-ND). These results highlight that the adaptive architecture selection ability of the \pinn\, is essential for PDE-aware architecture choice and cross-task generalization.

\paragraph{The Impact of Code Agent.}

\begin{figure}[!htb]
  \centering
  \begin{minipage}{0.4\textwidth}
    \centering
    \includegraphics[width=\linewidth]{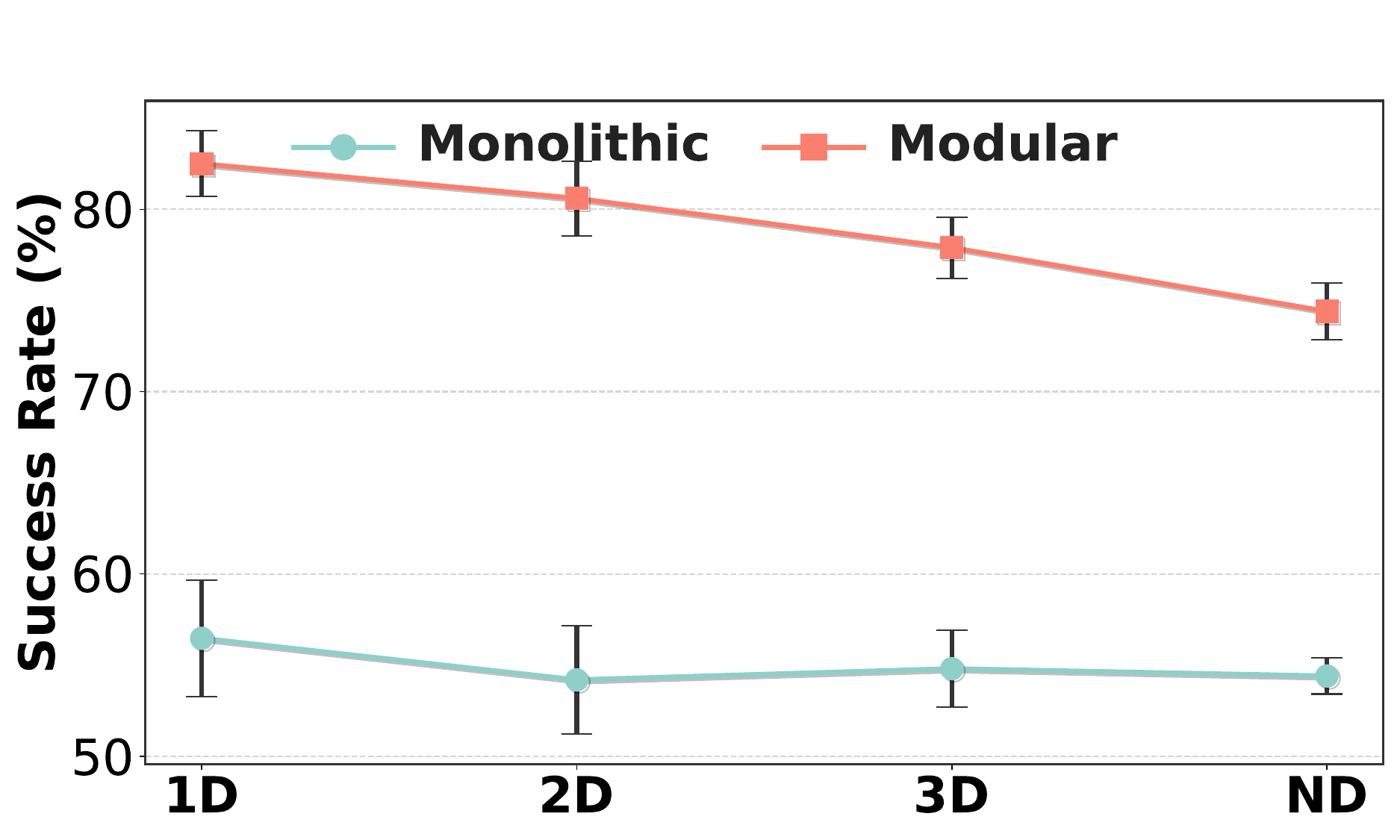}
    \caption{Ablation on the \emph{Code Agent}: success rate (\%)of monolithic vs.~modular code generation.}
    \label{fig:ab_code}
  \end{minipage}\hfill
  \begin{minipage}{0.58\textwidth}
    \centering
    \includegraphics[width=\linewidth]{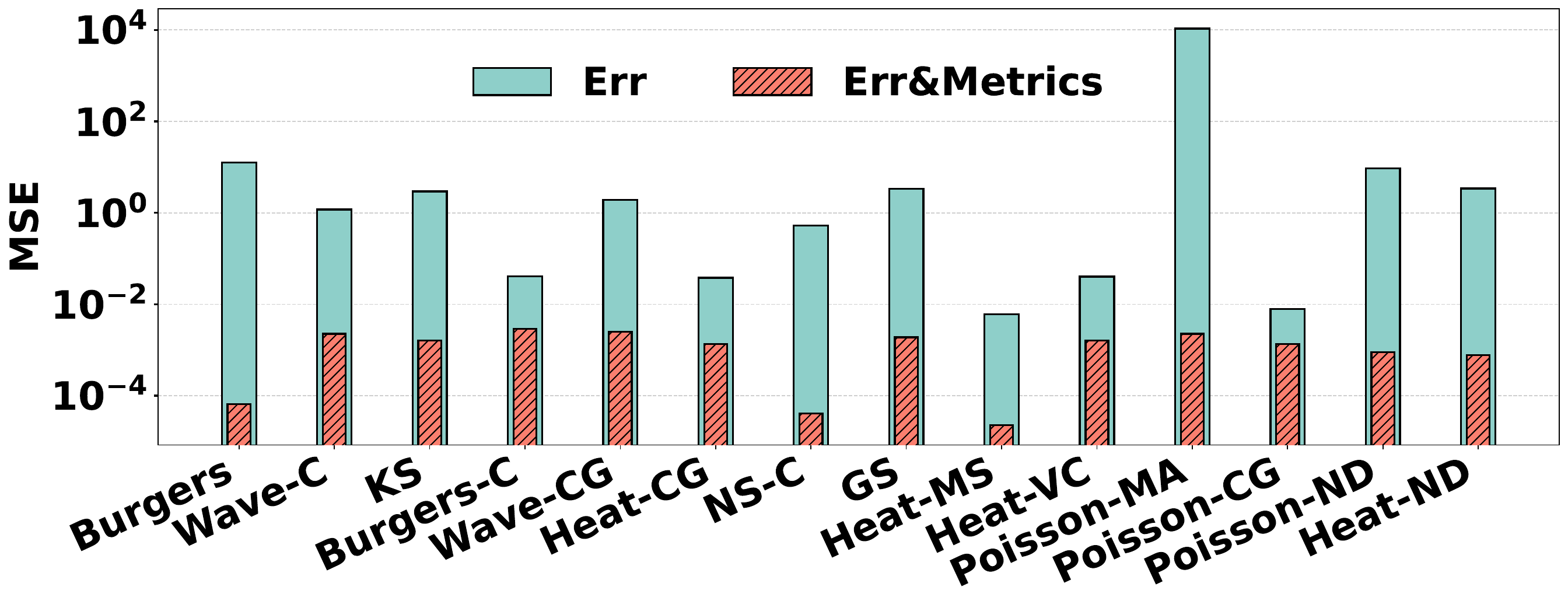}
    \caption{Ablation on the \emph{Feedback Agent}: MSE comparison of error-only feedback (\textit{Err})vs.~error feedback augmented with code quality metrics (\textit{Err\&Metrics}).}
    \label{fig:ab_feedback}
  \end{minipage}
\end{figure}

To validate the Impact of the \textbf{Code Agent}, we compare its modular code generation paradigm with a monolithic generator that attempts to produce the entire code in one pass. In the monolithic setting, runtime errors are hard to localize and every correction requires regenerating the full script, resulting in fragile execution. By contrast, the Code Agent decomposes the pipeline into modules (model, loss, training loop), allowing localized correction and reuse of valid components. As shown in Fig.~\ref{fig:ab_code}, this modular design improves the execution success rate by over 20\% across PDEs, highlighting the central role of the Code Agent in ensuring executability.

\paragraph{The Impact of Feedback Agent.}

We next evaluate the \textbf{Feedback Agent}, focusing on how different feedback signals affect the quality of the trained PINNs. The baseline uses only error messages from failed executions to guide refinement. Our full design augments these signals with the multi-dimensional quality metrics introduced in Sec~\ref{sec:Feedback-Agent}, including loss smoothness, gradient stability, and convergence behavior. As shown in Fig.~\ref{fig:ab_feedback}, the additional metrics consistently reduce MSE across PDE benchmarks, in some cases by several orders of magnitude. These results confirm that the Feedback Agent’s metric-guided feedback is crucial for achieving accuracy improvements once executability has been secured by the Code Agent.

\section{Conclusion}
\label{sec:conclusion}

We introduced \name, a multi-agent framework that constructs trainable physics-informed neural networks (PINNs)directly from natural-language task descriptions by integrating PDE parsing, architecture selection, modular code generation, and feedback refinement. Experiments on 14 PDEs show that \name\, achieves lower errors, higher execution success rates, and significantly reduced time overhead compared to strong baselines, while ablations confirm the value of modular generation, feedback-driven diagnostics, and knowledge-guided design. This work highlights the potential of LLM-based agents to bridge scientific intent and executable models, with future efforts focusing on multi-physics systems, irregular geometries, and noisy real-world data.

\clearpage
\section*{Ethics Statement}

This work does not involve human subjects, sensitive personal data, or experiments that could raise ethical concerns. The datasets used are publicly available, and no privacy or security issues are implicated. Our study focuses purely on methodological and computational aspects, and therefore we do not anticipate any direct ethical or societal risks arising from this research.

\section*{Reproducibility Statement}

We have made extensive efforts to ensure the reproducibility of our results. The descriptions of the proposed models and algorithms are included in the main text, while additional implementation details, hyperparameter settings, and training procedures are provided in the appendix and supplementary material. Information about datasets and data preprocessing steps is clearly documented. To further facilitate reproducibility, we provide an anonymous repository containing the source code, experiment scripts, and configuration files in the supplementary materials.

\section*{The Use of Large Language Models}

In preparing this manuscript, we used LLM to refine the clarity, fluency, and readability of the English writing. The LLM was employed only for linguistic polishing and expression improvement. All scientific content, analysis, results, and conclusions were conceived, validated, and written by the authors. The authors take full responsibility for the accuracy and integrity of the scientific claims presented in this paper.

\bibliography{custom}
\bibliographystyle{iclr2026_conference}

\clearpage
\section*{Appendix}
\setcounter{section}{0}
\renewcommand{\thesection}{\arabic{section}}

\setcounter{figure}{0}
\renewcommand{\thefigure}{\arabic{figure}}

\setcounter{table}{0}
\renewcommand{\thetable}{\arabic{table}}

\setcounter{equation}{0}
\renewcommand{\theequation}{\arabic{equation}}

\section{Details of Feature–Capability Encoding in the PINN Agent}
\label{app:pinnagent-metrics}

In Sec.~\ref{sec:PINN-Agent}, we introduced the idea of aligning PDE features with architecture capabilities through a weighted similarity score. This appendix provides the detailed definitions of both sides of the encoding.

\subsection{PDE Feature Representation}
Each PDE $E$ is mapped to a feature vector
\begin{equation}
\phi(E)= [f_{\mathrm{per}}(E),\, f_{\mathrm{geo}}(E),\, f_{\mathrm{ms}}(E)]^\top \in [0,1]^3,
\end{equation}
capturing three complementary aspects:

\paragraph{Periodicity.}
The degree of periodicity is quantified as
\begin{equation}
f_{\mathrm{per}}(E)= \frac{|\mathcal{P}(E)|}{d},
\end{equation}
where $d$ is the number of spatial dimensions and $\mathcal{P}(E)$ the set of periodic axes. Fully periodic domains yield $f_{\mathrm{per}}=1$, non-periodic domains yield $0$, and mixed cases take intermediate values.

\paragraph{Geometry complexity.}
We define
\begin{equation}
f_{\mathrm{geo}}(E)= \mathrm{clip}\!\left(\lambda_\Omega c_\Omega + \lambda_{\mathrm{disc}} c_{\mathrm{disc}},\,0,1\right),
\end{equation}
where $c_\Omega$ denotes domain irregularity ($0$ for rectilinear, $0.3$ for curved, $0.6$ for multi-component, $0.9$ for highly irregular)and $c_{\mathrm{disc}}$ denotes discretization irregularity ($0$ for Cartesian grids, $0.5$ for structured curvilinear, $0.8$ for unstructured FEM). The coefficients $\lambda_\Omega,\lambda_{\mathrm{disc}}$ control weighting.

\paragraph{Multi-scale demand.}
We measure the presence of multi-scale effects as
\begin{equation}
f_{\mathrm{ms}}(E)= \sigma\!\Big(
\alpha \cdot \mathbb{1}_{\{m \ge 3\}}
+ \beta \cdot \mathbb{1}_{\{\text{nonlinear}\}}
+ \gamma \cdot \log(1+\mathrm{Re/Pe})
+ \delta \cdot \mathbb{1}_{\{\text{nonlocal/fractional}\}}
\Big)\cdot \eta,
\end{equation}
where $m$ is the highest derivative order, $\mathrm{Re}$ and $\mathrm{Pe}$ are nondimensional numbers when available, and $\sigma(x)=(1+e^{-x})^{-1}$ normalizes values into $[0,1]$. $\eta=0.5$ attenuates trivial diffusion cases.

\subsection{PINN Architecture Capability Representation}
Each candidate architecture $\mathcal{A}$ is mapped to a capability vector
\begin{equation}
\psi(\mathcal{A})= [a_{\mathrm{per}}(\mathcal{A}),\, a_{\mathrm{geo}}(\mathcal{A}),\, a_{\mathrm{ms}}(\mathcal{A})]^\top \in [0,1]^3,
\end{equation}
indicating its inductive bias on the same three axes. The values are obtained from LLM reasoning informed by prior empirical studies, then normalized to $[0.1,0.9]$ to ensure comparability. For example:
\begin{itemize}
    \item Fourier-MLP: $(0.9,0.2,0.5)$ — strong on periodicity, weak on geometry, moderate on multi-scale demand.
    \item GNN: $(0.1,0.8,0.5)$ — strong on irregular geometry, moderate on multi-scale, weak on periodicity.
    \item Transformer: $(0.2,0.5,0.7)$ — strong on multi-scale via global attention, moderate on geometry, weak on periodicity.
    \item CNN: $(0.2,0.4,0.3)$ — effective on Cartesian grids, weak on irregular geometries and multi-scale.
    \item MLP: $(0.1,0.2,0.4)$ — generally applicable but with low inductive bias.
\end{itemize}


\section{Feedback Agent Quality Metrics}
\label{app:feedback-metrics}

The validation score produced by the \feedback\, agent aggregates four normalized metrics, each designed to capture a complementary aspect of code quality. Below we detail the first three metrics; the robustness metric is described separately.
\paragraph{(i)Convergence efficiency.} 
Convergence efficiency measures how quickly a model reaches a stable solution. We define it based on the number of training steps required for the loss to fall below a pre-specified tolerance~$\tau$:
\begin{equation}
T_{\text{conv}} = \min \{ t \mid L_t \leq \tau \},
\quad 
m_{\text{conv}} = \frac{1}{T_{\text{conv}}},
\end{equation}
where $L_t$ denotes the training loss at iteration $t$. A smaller $T_{\text{conv}}$ leads to a higher convergence score.  
For comparability across models, we normalize the score using the range of convergence steps observed in the search space:
\begin{equation}
\hat{m}_{\text{conv}} = 
\frac{T_{\max} - T_{\text{conv}}}{T_{\max} - T_{\min}},
\end{equation}
where $T_{\min}$ and $T_{\max}$ denote, respectively, the fastest and slowest convergence times among all candidates. This normalization ensures $\hat{m}_{\text{conv}} \in [0,1]$, with higher values indicating more efficient convergence.
\paragraph{(ii)Predictive accuracy.} 
Accuracy is assessed by the discrepancy between the model output and the governing PDE. Specifically, we compute the mean squared error (MSE)of the PDE residual over the training domain:
\begin{equation}
m_{\text{acc}} = - \mathrm{MSE}\big(\mathcal{N}_\theta, E \big),
\end{equation}
where $\mathcal{N}_\theta$ denotes the physics-informed neural network (PINN)parameterized by $\theta$, and $E$ represents the target PDE operator. The negative sign ensures that lower residual error corresponds to a higher accuracy score.

\paragraph{(iii)Model complexity.} 
Complexity reflects the resource demand of the model. We quantify it by the number of trainable parameters (or equivalently the computational cost in FLOPs), normalized with respect to the maximum within the search space:
\begin{equation}
m_{\text{comp}} = \frac{\#\mathrm{Params}(\mathcal{N}_\theta)}{\max \#\mathrm{Params}},
\end{equation}
where $\#\mathrm{Params}(\mathcal{N}_\theta)$ is the parameter count of the candidate PINN and $\max \#\mathrm{Params}$ is the maximum parameter count among all models considered. A lower value of $m_{\text{comp}}$ indicates a more compact architecture.

\paragraph{(iv)Robustness.} 
We quantify robustness by combining two complementary indicators. The first indicator, \emph{loss smoothness}, measures the stability of the training trajectory. Intuitively, when the loss fluctuates strongly across iterations, the optimization process is less reliable. We capture this by computing the normalized variation of the loss:
\begin{equation}
m_{\text{smooth}} = 1 - \frac{\mathrm{Std}(\Delta L_t)}{\mathrm{Mean}(L_t)}, 
\quad \Delta L_t = L_t - L_{t-1},
\end{equation}
where $L_t$ denotes the training loss at iteration $t$, and $\Delta L_t$ is the difference between consecutive iterations. A higher value of $m_{\text{smooth}}$ indicates a smoother and more stable training curve.  

The second indicator, \emph{gradient health}, evaluates whether the gradient magnitude remains within a reasonable range, avoiding both vanishing and exploding gradients. Specifically,
\begin{equation}
m_{\text{grad}} =
\begin{cases}
1, & \epsilon \leq \dfrac{\|\nabla_\theta L\|}{d} \leq \kappa, \\
0, & \text{otherwise},
\end{cases}
\end{equation}
where $\nabla_\theta L$ is the gradient of the loss with respect to the parameters, $d$ is the number of parameters, and $\epsilon, \kappa > 0$ are user-defined thresholds specifying the acceptable lower and upper bounds of the normalized gradient magnitude.  

Finally, we define the robustness score as a convex combination of the two indicators:
\begin{equation}
m_{\text{rob}} = \alpha \, m_{\text{smooth}} + (1-\alpha)\, m_{\text{grad}},
\end{equation}
where $\alpha \in [0,1]$ is a weighting factor that balances the contributions of loss smoothness and gradient health. This formulation ensures that robustness reflects both stable optimization dynamics and well-conditioned gradients.

The overall validation score is defined as a weighted combination of the four normalized metrics:
\begin{equation}
S(C)= w_1 \, \hat{m}_{\text{conv}} 
     + w_2 \, \hat{m}_{\text{acc}} 
     + w_3 \, \hat{m}_{\text{comp}} 
     + w_4 \, \hat{m}_{\text{rob}},
\end{equation}
where $w_1, w_2, w_3, w_4 \geq 0$ are user-specified weights that control the relative importance of convergence efficiency, predictive accuracy, model complexity, and robustness, respectively. By tuning the weights, one can emphasize different aspects of model quality depending on the application

\section{Task2PDE Dataset}\label{app:task2pde}

To rigorously evaluate the ability of language models to map natural-language task descriptions into formal PDE specifications, we construct the \textbf{Task2PDE} dataset. The dataset is derived from eight representative PDE families selected from the PINNacle benchmark \citep{hao2023pinnacle}, spanning different spatial dimensions:

\begin{itemize}
    \item \textbf{1D:} Burgers’, Wave–C, Kuramoto–Sivashinsky (KS);
    \item \textbf{2D:} Heat–MS, Poisson–MA, incompressible Navier–Stokes (NS–C);
    \item \textbf{3D:} Poisson–CG;
    \item \textbf{High-dimensional ND:} Heat–ND.
\end{itemize}

For each PDE family, we construct $50$ distinct task descriptions under four difficulty levels, yielding a total of $8 \times 4 \times 50 = 1600$ samples. Each sample is paired with its ground-truth PDE formulation, including operators, coefficients, boundary/initial conditions, and domain specification. This ensures that every natural-language description corresponds uniquely to one PDE instance, enabling systematic evaluation of semantic-to-symbol grounding.

\paragraph{Four Levels of Description.}
We design four difficulty levels to simulate progressively more challenging natural-language inputs. The same PDE is described at each level, but the linguistic form becomes increasingly noisy, redundant, and disordered. Below we illustrate the differences using a 2D heat diffusion example (a square plate with mixed boundary conditions).

\vspace{0.5em}
\noindent\textbf{Level 1 — Clean.}  
\emph{A thin, square metal plate is placed horizontally on an insulated table, with all four edges exposed to the surrounding air. The plate’s initial temperature distribution is given as a spatially varying function. During the experiment, two opposite edges are kept at distinct, constant temperatures, while the remaining two edges are perfectly insulated. As time progresses, the temperature evolves by heat diffusion and eventually reaches a steady state.}  
\newline
\textit{Characteristics:} concise, physics-only, no irrelevant content.

\vspace{0.5em}
\noindent\textbf{Level 2 — With Nonsense.}  
\emph{A thin, square metal plate is placed horizontally on an insulated table in a laboratory (the lab’s coffee machine was unusually loud today). The plate’s initial temperature distribution is established through a heating process (which the technician jokingly described as “painting with heat”). Two opposite edges are kept at constant but different temperatures, while the other two edges are insulated. The ambient air has negligible effect (ignoring occasional drafts from the door). As time progresses, the plate’s temperature diffuses toward a steady state.}  
\newline
\textit{Characteristics:} adds irrelevant noise (coffee machine, jokes, drafts), while keeping the physics intact.

\vspace{0.5em}
\noindent\textbf{Level 3 — Redundant Rephrasing.}  
\emph{A thin, square metal plate is placed horizontally on an insulated table (the coffee machine was loud today). Its initial temperature distribution is established through a heating process (the technician called it “painting with heat”). Two opposite edges are kept at constant temperatures — that is, one side fixed hot, the other cooler. The other two edges are insulated — in other words, no heat flux, meaning the normal derivative of temperature is zero. The ambient air is negligible (equivalently, convective exchange is disregarded). As time progresses, heat diffuses and the plate approaches steady state, i.e., the time derivative eventually vanishes.}  
\newline
\textit{Characteristics:} retains Level~2 noise, adds redundant reformulations of the same physics.

\vspace{0.5em}
\noindent\textbf{Level 4 — Disordered Bullet Style.}  
\emph{A thin, square metal plate is placed horizontally on an insulated table — eventually the temperature approaches a steady state; the ambient air is negligible (ignoring drafts from the door); the initial distribution is set by a heating process (“painting with heat”); two opposite edges are kept at constant temperatures (one hot, one cool); as time evolves, heat diffuses across the plate; all four edges are exposed to air; the remaining two edges are insulated (no flux, i.e., normal derivative zero).}  
\newline
\textit{Characteristics:} retains noise and redundancy, but breaks logical order into fragmented, pseudo-bullet sentences.

\paragraph{Purpose.}
By varying linguistic complexity in controlled steps, Task2PDE moves beyond benchmarks that assume formal PDE input. The four-level design enables fine-grained evaluation of whether models can (i)ignore irrelevant information, (ii)consolidate redundant rephrasings, and (iii)reconstruct structured PDE specifications from fragmented input. Combined with eight PDE families spanning 1D to high-dimensional settings, the dataset provides a comprehensive testbed for evaluating natural-language-driven PDE solvers such as \name.

\section{Extended Results: MSE and Success Rate across PDE Benchmarks}\label{app:complete_main}

For completeness, we report the full experimental results across all $14$ PDE benchmarks. 
Table~\ref{tab:app_results_fixed} presents the mean squared error (MSE)together with standard deviations, complementing the aggregated results in the main text. 
Table~\ref{tab:main_results} provides per-PDE success rates (\%)averaged over $10$ runs, offering a more fine-grained view of performance across different equations and dimensions. 

These results serve as a detailed supplement to the main comparisons: 
our method consistently achieves the lowest average errors with significantly reduced variance, and obtains higher success rates across nearly all PDEs. 
In particular, \name\, improves code executability and training stability even for challenging high-dimensional and chaotic cases, reinforcing the conclusions drawn in the main paper.

\renewcommand{\arraystretch}{1.4}
\setlength{\tabcolsep}{12pt}

\begin{table*}[t]
\centering
\caption{Comparative performance (MSE)of \name\, and baseline approaches on 14 different PDEs. Results are averaged over 10 runs.}
\label{tab:app_results_fixed}
\resizebox{\textwidth}{!}{
\begin{tabular}{lccccccc}
\hline
\textbf{PDEs} & \textbf{RandomAgent} & \textbf{BayesianAgent} & \textbf{PINNsAgent} & \textbf{PINNacle} & \textbf{SCoT} & \textbf{Self-Debug} & \textbf{Ours}\\
\hline
\multicolumn{8}{c}{\textbf{1D}} \\
\hline
Burgers & 6.63E-02 ($\pm$1.10E-01)& 8.70E-02 ($\pm$6.51E-03)& 1.10E-04 ($\pm$7.76E-05)& {\color{gray}7.90E-05} & 1.40E+01 ($\pm$1.06E+00)& 1.26E+01 ($\pm$9.54E-01)& 6.48E-05 ($\pm$9.00E-05)\\
Wave-C & 1.50E-01 ($\pm$1.46E-01)& 1.78E-01 ($\pm$3.84E-02)& 3.74E-02 ($\pm$4.32E-02)& {\color{gray}3.01E-03} & 1.28E+00 ($\pm$6.21E-02)& 1.18E+00 ($\pm$5.72E-02)& 2.25E-03 ($\pm$1.80E-04)\\
KS & 1.09E+00 ($\pm$3.58E-02)& 1.10E+00 ($\pm$2.55E-03)& 1.09E+00 ($\pm$3.20E-02)& {\color{gray}1.04E+00} & 3.33E+00 ($\pm$7.80E-02)& 2.93E+00 ($\pm$6.86E-02)& 1.62E-03 ($\pm$1.35E-04)\\
\hline
\multicolumn{8}{c}{\textbf{2D}} \\
\hline
Burgers-C & 2.48E-01 ($\pm$4.04E-03)& 2.42E-01 ($\pm$8.96E-03)& 2.93E-01 ($\pm$2.43E-02)& {\color{gray}1.09E-01} & 4.54E-01 ($\pm$5.57E-02)& 4.09E-02 ($\pm$5.01E-03)& 2.88E-03 ($\pm$2.25E-04)\\
Wave-CG & 2.87E-02 ($\pm$4.98E-04)& 2.11E-02 ($\pm$1.12E-02)& 4.59E-02 ($\pm$1.68E-02)& {\color{gray}2.99E-02} & 2.00E+00 ($\pm$1.62E-01)& 1.90E+00 ($\pm$1.54E-01)& 2.52E-03 ($\pm$1.62E-04)\\
Heat-CG & 3.96E-01 ($\pm$3.22E-01)& 1.17E-01 ($\pm$3.24E-02)& 9.06E-02 ($\pm$2.69E-01)& {\color{gray}8.53E-04} & 4.38E+00 ($\pm$3.48E-01)& 3.81E-02 ($\pm$3.03E-03)& 1.35E-03 ($\pm$9.00E-05)\\
NS-C & 4.02E-03 ($\pm$5.93E-03)& 5.12E-03 ($\pm$1.33E-03)& 1.40E-05 ($\pm$1.12E-05)& {\color{gray}2.33E-05} & 5.67E-01 ($\pm$6.28E-02)& 5.27E-01 ($\pm$5.84E-02)& 4.05E-05 ($\pm$4.50E-05)\\
GS & 4.28E-03 ($\pm$2.23E-05)& 4.03E-03 ($\pm$4.47E-04)& 3.37E+08 ($\pm$1.01E+09)& {\color{gray}4.32E-03} & 3.76E+00 ($\pm$5.27E-02)& 3.35E+00 ($\pm$4.69E-02)& 1.89E-03 ($\pm$1.44E-04)\\
Heat-MS & 1.84E-02 ($\pm$1.18E-02)& 7.48E-03 ($\pm$3.81E-03)& 1.06E-04 ($\pm$1.86E-04)& {\color{gray}5.27E-05} & 7.10E-02 ($\pm$3.05E-03)& 6.04E-03 ($\pm$2.59E-04)& 2.27E-05 ($\pm$7.20E-05)\\
Heat-VC & 3.57E-02 ($\pm$8.72E-03)& 3.93E-02 ($\pm$2.17E-03)& 1.43E-02 ($\pm$1.77E-02)& {\color{gray}1.76E-03} & 4.46E+00 ($\pm$1.05E+00)& 4.01E-02 ($\pm$9.45E-03)& 1.62E-03 ($\pm$1.08E-04)\\
Poisson-MA & 5.87E+00 ($\pm$1.17E+00)& 5.82E+00 ($\pm$2.30E+00)& 3.16E+00 ($\pm$9.92E-01)& {\color{gray}1.83E+00} & 1.24E+04 ($\pm$5.71E+03)& 1.07E+04 ($\pm$4.91E+03)& 2.25E-03 ($\pm$1.35E-04)\\
\hline
\multicolumn{8}{c}{\textbf{3D}} \\
\hline
Poisson-CG & 3.82E-02 ($\pm$2.15E-02)& 2.55E-02 ($\pm$5.65E-03)& 3.35E-02 ($\pm$2.18E-02)& {\color{gray}9.51E-04} & 4.17E-02($\pm$3.77e-03)& 9.51E-03($\pm$1.35e-03)& 1.35E-03 ($\pm$9.00E-05)\\
\hline
\multicolumn{8}{c}{\textbf{ND}} \\
\hline
Poisson-ND & 1.30E-04 ($\pm$2.78E-04)& 4.72E-05 ($\pm$2.76E-06)& 4.77E-04 ($\pm$3.21E-05)& {\color{gray}2.09E-06} & 9.93E+00 ($\pm$6.51E-03)& 9.43E+00 ($\pm$6.18E-03)& 842.00E-06 ($\pm$5.17E-07)\\
Heat-ND & 2.58E-00 ($\pm$9.87E-02)& 1.18E-04 ($\pm$8.92E-06)& 8.57E-04 ($\pm$1.31E-06)& {\color{gray}8.52E+00} & 3.74E+00 ($\pm$3.29E-01)& 3.40E-03 ($\pm$2.99E-04)& 4.72E-04($\pm$6.30E-05)\\
\hline
\end{tabular}
}
\end{table*}

\renewcommand{\arraystretch}{1.4}
\setlength{\tabcolsep}{12pt}

\begin{table*}[t]
\centering
\caption{Success rate (\%)of \name\, and baseline approaches on 14 different PDEs. Results are averaged over 10 runs.}
\label{tab:main_results}
\resizebox{\textwidth}{!}{
\begin{tabular}{lcccccc}
\hline
\textbf{PDEs} & \textbf{RandomAgent} & \textbf{PINNsAgent} & \textbf{PINNacle }& \textbf{SCoT} & \textbf{Self-Debug} & \textbf{Ours} \\
\hline
\multicolumn{7}{c}{\textbf{1D}} \\
\hline
Burgers      & 29.7\% & 36.2\% & {\color{gray}38.9\%} & 58.6\% & 59.7\% & 84.3\% \\
Wave-C       & 28.5\% & 34.8\% & {\color{gray}37.2\%} & 57.2\% & 58.3\% & 80.7\% \\
KS           & 27.9\% & 33.5\% & {\color{gray}35.9\%} & 55.1\% & 56.4\% & 82.5\% \\
\hline
\multicolumn{7}{c}{\textbf{2D}} \\
\hline
Burgers-C    & 26.1\% & 33.4\% & {\color{gray}36.2\%} & 56.3\% & 58.0\% & 81.1\% \\
Wave-CG      & 25.4\% & 31.2\% & {\color{gray}34.0\%} & 54.9\% & 56.1\% & 77.4\% \\
Heat-CG      & 25.1\% & 32.6\% & {\color{gray}35.1\%} & 55.7\% & 57.0\% & 81.6\% \\
NS-C         & 26.3\% & 34.1\% & {\color{gray}36.8\%} & 57.1\% & 58.9\% & 83.3\% \\
GS           & 24.9\% & 30.7\% & {\color{gray}33.2\%} & 53.8\% & 55.0\% & 78.8\% \\
Heat-MS      & 26.8\% & 35.0\% & {\color{gray}37.6\%} & 58.4\% & 59.6\% & 82.7\% \\
Heat-VC      & 25.6\% & 32.0\% & {\color{gray}34.5\%} & 55.2\% & 56.8\% & 80.5\% \\
Poisson-MA   & 23.7\% & 29.8\% & {\color{gray}32.7\%} & 52.7\% & 54.1\% & 79.2\% \\
\hline
\multicolumn{7}{c}{\textbf{3D}} \\
\hline
Poisson-CG   & 22.9\% & 30.4\% & {\color{gray}33.5\%} & 53.2\% & 54.8\% & 77.9\% \\
\hline
\multicolumn{7}{c}{\textbf{ND}} \\
\hline
Poisson-ND   & 21.7\% & 28.9\% & {\color{gray}31.7\%} & 51.5\% & 53.1\% & 73.3\% \\
Heat-ND      & 20.9\% & 29.6\% & {\color{gray}32.4\%} & 52.1\% & 53.7\% & 75.5\% \\
\hline
\end{tabular}}
\end{table*}



\end{document}

%% file: math_commands.tex
\usepackage{amsmath,amsfonts,bm}

\def\eqref#1{equation~\ref{#1}}

\def\1{\bm{1}}

\DeclareMathAlphabet{\mathsfit}{\encodingdefault}{\sfdefault}{m}{sl}
\SetMathAlphabet{\mathsfit}{bold}{\encodingdefault}{\sfdefault}{bx}{n}

